# Visualizing Machine Learning Models for Enhanced Financial Decision-Making and Risk Management

Priyam Ganguly
*Hanwha Q CELLS America Inc*
Irvine, California, USA
priyam.develop@gmail.com

Ramakrishna Garine
*University of North Texas*
Denton ,Texas, USA
ramakrishnagarine@gmail.com

Isha Mukherjee
*Pace University*
New York, USA
ishamukherjee123@gmail.com

*Abstract*— **This study emphasizes how crucial it is to visualize machine learning models, especially for the banking industry, in order to improve interpretability and support predictions in high-stakes financial settings. Visual tools enable performance improvements and support the creation of innovative financial models by offering crucial insights into the algorithmic decision-making processes. Within a financial machine learning framework, the research uses visually guided experiments to make important concepts, such risk assessment and portfolio allocation, more understandable. The study also examines variations in trading tactics and how they relate to risk appetite, coming to the conclusion that the frequency of portfolio rebalancing is negatively correlated with risk tolerance. Finding these ideas is made possible in large part by visualization. The study concludes by presenting a novel method of locally stochastic asset weighing, where visualization facilitates data extraction and validation. This highlights the usefulness of these methods in furthering the field of financial machine learning research.**

## I. Introduction

The rapid expansion of the Machine Learning (ML) sector has resulted in the development of highly sophisticated algorithms with immense computational capabilities, many of which have grown too complex for human comprehension [1], [2], [3], [4], [5]. This increasing complexity has given rise to a lack of interpretability, which raises significant ethical concerns regarding the transparency of the decision-making processes in these algorithms. In high-stakes environments such as medicine and law, where ML models are employed, it is crucial to ensure that the individuals making decisions based on these models' predictions are able to understand the underlying reasoning. For instance, Artificial Neural Networks (ANNs) and Natural Language Processing (NLP) algorithms are frequently utilized to derive insights from unstructured medical data. In these cases, the ability to comprehend the motivations behind the models' decisions is paramount. Therefore, visualization techniques can help elucidate these decision-making processes, thereby enhancing trust and ensuring the reliability of the predictions, particularly in fields where the consequences of errors are significant. Another critical aspect of developing ML models is evaluating their performance in real-world applications. Often, models behave differently when tested on real-world data as compared to their performance on metrics of interest during the training phase. This discrepancy arises because real-world data can deviate significantly from emulated data, resulting in unexpected model behavior after deployment. Through visualization, developers gain a deeper understanding of the internal dynamics of these models, allowing for necessary modifications to be made in order to bridge the performance gap between training and deployment. This approach not only improves the reliability of the model but also maintains its overall fidelity.

Therefore, in response to the challenges associated with the "black box" nature of traditional ML models, the Tsetlin Machine (TM) has emerged as a promising solution. The TM uses propositional logic and supervised learning to formulate clauses, making the decision-making process more interpretable compared to traditional neural networks. The clarity offered by the TM's architecture allows for greater insight into how decisions are made, positioning it as a valuable alternative for domains where transparency is essential. This study focuses on visualizing the decision-making processes of the TM and analyzing the internal dynamics, particularly the average number of Tsetlin Automaton (TA) state flips after adjusting key parameters during initialization. Furthermore, the study explores the withdrawal of certain clauses in cases where significant differences in clause performance are observed, with the goal of reducing computational load while preserving optimal accuracy such that the first objective is to acquire a comprehensive understanding of the TM's architecture, particularly focusing on clause computation and class voting. In addition, the study will break down the clause computation logic outside of the program's libraries to enhance clarity and understanding and focuses on analyzing the average number of TA state flips (i.e., transitions between include and exclude states) across all 10 classes in the MNIST dataset. The study will extract data on TA flips from the TM during training and develop a script to calculate and plot the average number of flips for different values of the 's' and 'T' hyperparameters. By adjusting these parameters, the study aims to explore how they influence the TM's learning dynamics.
The study is as follows; similar papers are shown in the following section. The methodology is explained in Section III. The experimental analysis is carried out in Section IV. We wrap up the paper with some conclusions and ideas for further research in Section V.

## II. RELATED WORKS

ANNs are widely acknowledged for their foundational role in advancing ML and shaping the future of computational intelligence such as [6], [7], [8]. Much like the human brain, ANNs utilize neurons as the basic computational units, processing inputs weighted according to their significance to detect patterns. However, recent developments in the TM, a low-complexity ML model, have demonstrated that it can effectively compete with ANNs, particularly in areas such as learning convergence [9], [10], [11]. Notably, the TM boasts up to 15 times greater energy efficiency compared to ANNs, though this efficiency comes with an increase in energy consumption per epoch. Unlike ANNs, which often operate as opaque "black-box" models lacking transparency, the TM's rule-based algorithm offers inherent interpretability, allowing for clearer insights into its decision-making process. This transparency contributes to a higher degree of reliability, making TMs particularly suitable for deployment in high-stakes environments such as healthcare and legal systems, where trust in ML models is essential. A recurring challenge in ML is the trade-off between accuracy and interpretability such as [12], [13], [14]. While high accuracy may suffice in less consequential applications, complex problems often require insight into the model's decision-making process, even if this compromises precision. Understanding the rationale behind the predictions enables researchers to detect potential biases and adjust the model to improve its overall performance. The importance of interpretability in ML has led to two main approaches [15]: model-based intrinsic interpretability, which offers real-time insight during the model's operation, and post-hoc extrinsic interpretability, which provides analysis after the model has completed its task. This study seeks to develop visualization tools that encompass both approaches, enhancing transparency and understanding of the TM's inner workings.

The TM's interpretability is derived from its ability to construct highly discriminative clauses in propositional logic, forming patterns in the data that facilitate human inference. Several modifications to the TM's learning mechanisms have been proposed to further enhance its clause discrimination power such as [16], [17], [18]. One such improvement is the Weighted Tsetlin Machine (WTM), which assigns weights to clauses, thus reducing both computation time and memory usage. Building on this, the Integer Weighted TM (IWTM) was developed to address the accuracy-interpretability trade-off. The IWTM achieves higher accuracy by grouping weaker clauses, allowing more effective clauses to function autonomously. This approach significantly reduces the number of literals required, thus improving both accuracy and interpretability. However, despite these advancements, no real-time visualizations were developed to track the formation of clauses during training, which represents a potential avenue for further exploration within this study. Another challenge in the current TM architecture is the issue of overfitting, where clauses capture irrelevant patterns or noise, diminishing the model's performance. However, this study also explores the concept of local clause removal as a method to improve the TM's performance and interpretability. By analyzing clause outputs in the correctly predicted class, underperforming clauses can be identified and randomly removed. This technique, referred to as locally random clause removal, targets the least useful clauses, enhancing the model's overall efficiency. The method builds on the stochastic clause dropping approach, but introduces a more targeted, yet still randomized, method of clause removal. This refinement aims to further improve both the accuracy and interpretability of the TM. Furthermore, the study seeks to develop visualization tools that provide real-time insights into the TM's decision-making processes, allowing researchers to track clause formation and state transitions throughout training.

## III. METHODOLOGY

The TM is designed to process Boolean input vectors. The corresponding set of literals where each feature may appear in its original or negated form. For example, if the input vector is $X = [0,1,1,0]$, the literal set becomes $\mathscr{L} = \{0,1,1,0,1,0,0,1\}$, which reflects both the literals and their negations. The TA, a fundamental component of the TM, addresses the multi-armed bandit problem by offering two possible actions: inclusion or exclusion. The objective is to dynamically allocate literals and negated literals to states that maximize the likelihood of receiving rewards. Literals that do not contribute positively to the outcome are demoted to states where they are excluded from the clause, thus achieving a reward. When a state is rewarded, the confidence in that state increases, while a penalty action diminishes the confidence, pushing the TA towards a state boundary. Penalties also prompt the TA to switch actions. Each literal is associated with its own TA, which determines its relevance to the clause. If a literal is included, it contributes to the clause, and each clause contains a set of TAs, one for each literal. This fixed finite-state approach is optimized for stochastic environments and features low computational complexity. Given the simple increment and decrement functions, the TA exhibits a minimal memory footprint, ensuring computational efficiency. Literals deemed relevant by the TA are combined using a conjunction operator (AND). The resulting clause is formulated by ANDing both included literals and negated literals, producing the general form for clause. For instance, given the input $X = [0,1,1,0]$ (Noisy XOR), indexed as $X = [x_o, x_1, x_2, x_3]$, the TM may generate a clause such as $C_k = x_o \wedge \neg x_1$. When evaluated, this clause yields: $C_k = 0 \wedge 1 = 0$. This demonstrates how the TM constructs clauses by combining relevant literals and negated literals. The clause cycle, beginning with the input data and culminating in the final clause, is visually represented in Fig. 1.

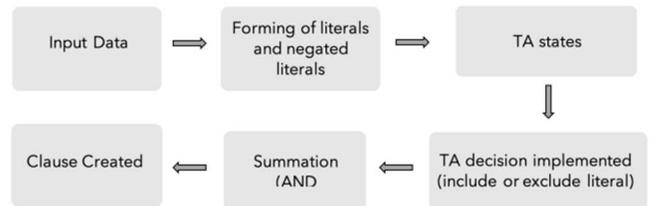

Fig. 1. A flow diagram illustrating how clauses are created inside the TM

Upon generating the clauses, the TM proceeds to vote on the class label most appropriate for the given input. In the case of the Noisy XOR dataset, which consists of two Boolean classes, the TM assigns clauses to one of the two classes, dividing them into positive and negative halves. Typically, clauses with odd indices are assigned negative polarity, while those with even indices are assigned positive polarity. For instance, in the case of Class $Y = 0$ for the Noisy XOR dataset, the following clauses may be generated:

1. Positive Polarity Clauses: Clause #0: $x_o \land x_1 = 0 \land 1 = 0$, and Clause #2: $x_o \land x_1 = 0 \land 1 = 0$.
2. Negative Polarity Clauses: Clause #1: $x_1 \land \neg x_o = 1 \land 1 = 1$, and Clause #3: $x_1 \land \neg x_o = 1 \land 1 = 1$.

The polarity of each clause determines whether it contributes positively (votes for) or negatively (votes against) the class prediction. The final decision is made by summing the votes from both positive and negative clauses and applying a threshold. After establishing clause polarities, the next step involves summing the outputs from both positive and negative clauses. The class confidence score, denoted as '$u$' is computed via $u = \Sigma(Positive\ Clause\ Values) + \Sigma(Negative\ Clause\ Values)$. For the Noisy XOR example, the negative clauses sum to a value of 2, which is then negated due to their negative polarity, yielding -2. The positive clauses sum to 0. The overall class confidence value thus becomes $u = 0 + (-2) = -2$. The final class prediction is determined through thresholding: if $u < 0$, the TM votes against the class; if $u \geq 0$, the TM votes in favor of the class. In this case, since $u = -2$, the TM votes against Class $Y = 0$, thereby implicitly voting for Class $Y = 1$. This prediction is correct, as the XOR of the first two input elements (0 and 1) is 1. However, to implement the inclusion or exclusion of literals in a clause, the TM applies Boolean algebra. Each literal is connected to its respective TA, where the action states of the TA (include or exclude) are pre-determined and represented as 1 (include) or 0 (exclude). Negated literals are handled in a similar manner by their corresponding TAs. The inclusion or exclusion of each literal is determined through Boolean operations, specifically ANDing ($\land$) and ORing ($\lor$) the literal with the negated TA state. For instance, consider two literals ($x_i$ and $x_s$) and their corresponding negated literals. This approach ensures that the combination of the literal and its associated TA state influences the final clause result. If any included literal-TA combination produces a 0, the entire clause value becomes 0.

## IV. EXPERIMENTAL ANALYSIS

The primary objective of this study was to develop a tool capable of training a TM on the Noisy XOR dataset, generating clauses from the training process, and visualizing the clause-based decision-making process on new, unseen input data. The study aimed to explore the interpretability of the TM, particularly how clause voting affects class predictions. Additionally, the study examined the effect of hyperparameters such as learning sensitivity and thresholding on the performance of the TM and proposed a novel method to enhance the TM architecture by optimizing clause efficiency. To achieve these goals, the TM Interpretability Tool was utilized to train the TM on the Noisy XOR dataset. This dataset consisted of 5000 inputs, each with 12 features, under a signal-to-noise ratio of 69.8%. The TM model used 10 clauses per class, divided into positive and negative polarities, with even-indexed clauses assigned positive polarity and odd-indexed clauses assigned negative polarity. The TM achieved 100% accuracy on the Noisy XOR dataset, demonstrating its capacity to handle this simple dataset effectively. After the TM was trained, the study focused on visualizing the interaction between clauses generated during training and new, unseen inputs. A Python script was developed to parse the clauses and cross-check the literals against the input vector provided by the user. The tool calculated the contributions of both positively and negatively polarized clauses and visualized the results on a graph. This graph displayed positive and negative class votes, where a positive bar indicated a vote for a class, and a negative bar signaled a vote against a class. The initial experimental results (Figs. 2 and 3) illustrated the TM's voting behavior when presented with unseen input data. For two specific input arrays, [1,0,1,1,0,0,0,0,0,1,0] and [1,1,1,1,0,0,0,0,0,1,0], the tool successfully visualized the TM's voting results. In the first case, the TM voted against class $Y = 0$ and in favor of class $Y = 1$, while in the second case, the TM voted for class $Y = 0$. These results confirmed that the clause summation and class voting processes were correctly implemented, fulfilling the project's objectives. This visualization was instrumental in shedding light on the TM's decision-making process, highlighting how each clause contributed to the final class vote. It also provided insights into the confidence levels of class predictions, with higher confidence levels indicated by larger class sums.

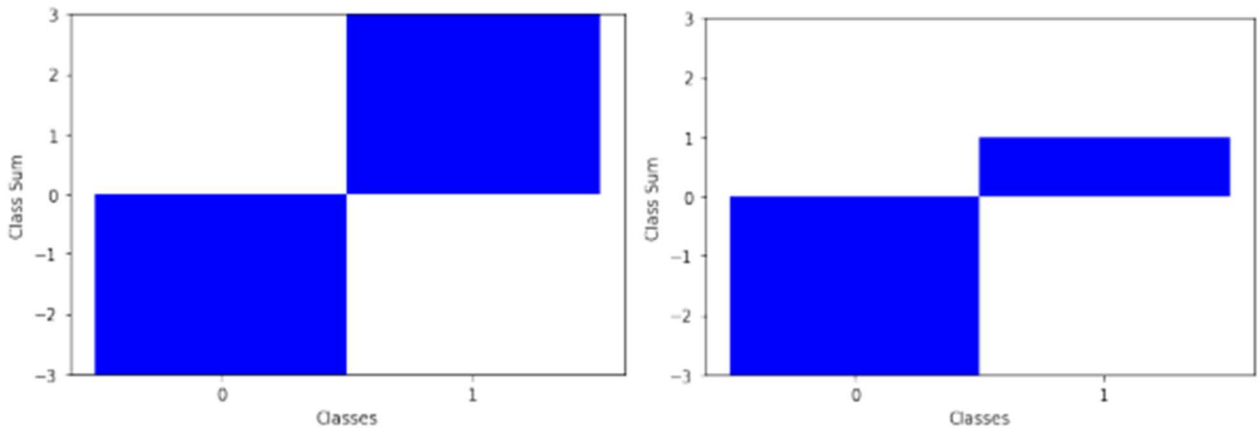

Fig. 2. Outcomes of the voting for Class 1 and against Class 0 using the visualization tool

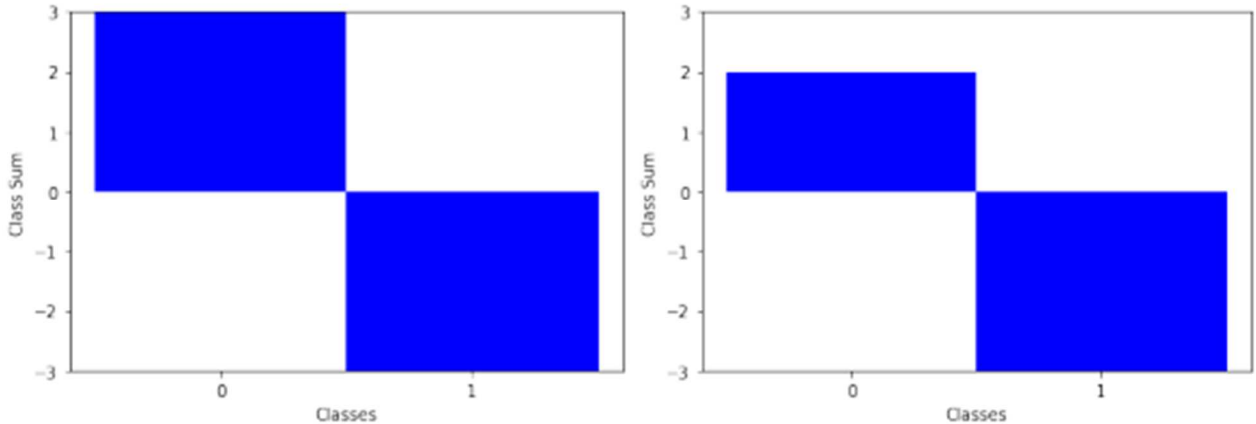
Fig. 3. Outcomes of the voting for Class 2 and against Class 3 using the visualization tool

A secondary goal of the study was to analyze the effects of varying TM hyperparameters on the number of TA flips and the overall accuracy of the TM. Specifically, two key hyperparameters were examined: learning sensitivity (denoted as 's') and thresholding (denoted as 'T'). In the first experiment, the sensitivity parameter 's' was varied while keeping the threshold value 'T' constant. The values of 's' ranged from 2.0 to 10.0, in increments of 2. For each value of 's', the average number of TA flips (ANOF) was plotted over time as shown in Fig. 4. As expected, the ANOF was initially high for all values of 's', due to the random assignment of TAs to middle states at the beginning of training. However, as the TM trained, the number of flips decreased as the TAs became more confident in their final states. It was observed that lower values of 's', particularly s = 4.0, resulted in significantly lower ANOF compared to higher values. This suggests that lower sensitivity biases the TAs toward excluding literals more frequently, thereby requiring fewer state transitions during training. Conversely, higher values of 's' increased the probability of including literals in the clauses, leading to a higher ANOF. Additionally, the rate of change in ANOF was found to be inversely proportional to the value of 's', indicating that as 's' increased, the TAs became more stable, and the flickering between include and exclude states diminished. In terms of accuracy, the TM performed best at s = 4.0, achieving an accuracy of 90.9%. As 's' increased beyond this value, the accuracy declined, suggesting that there is an optimal range of sensitivity values for maximizing accuracy while minimizing TA flips.

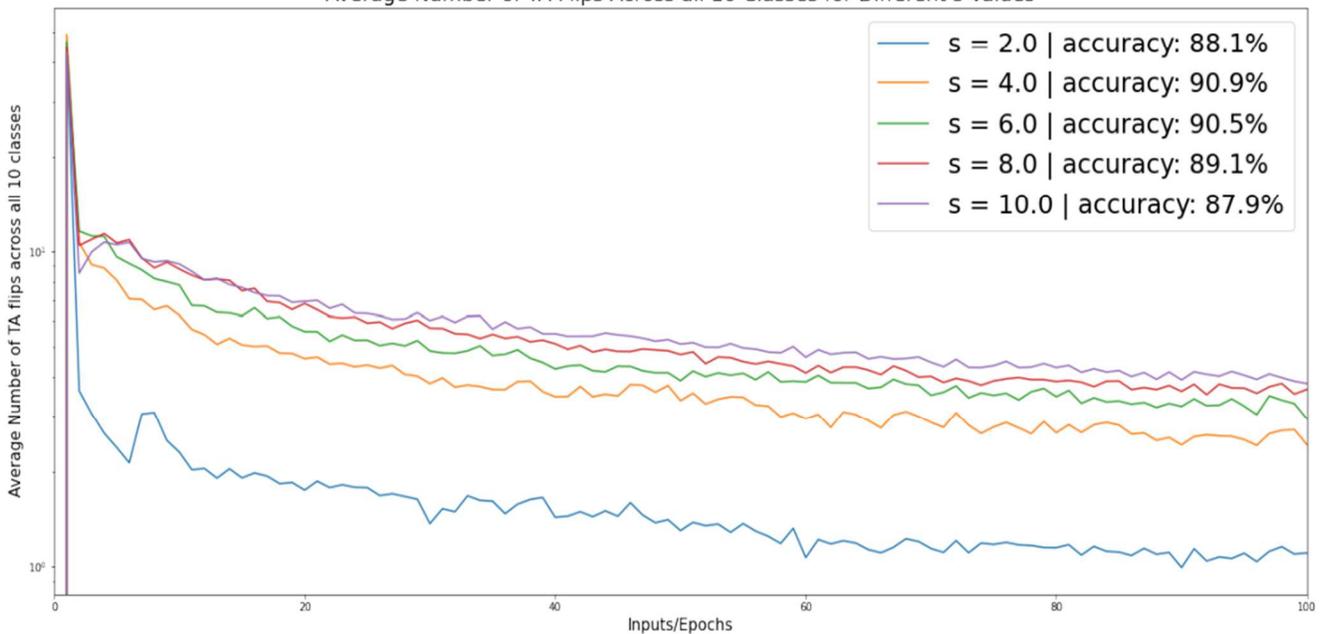
Fig. 4. The typical amount of TA flips for each of the ten classes for various values of 's'

The second part of the experiment involved varying the threshold value 'T' from 12.5 to 22.5, while keeping the sensitivity constant at s = 3.9 as shown in Fig. 5. Similar to the first experiment, the ANOF decreased as 'T' increased, since a higher threshold reduced the frequency of TA feedback, resulting in fewer flips. However, unlike the sensitivity experiment, accuracy remained relatively constant across different values of 'T', averaging around 90.72%. This could be attributed to the fact that the TM was only trained for three epochs, a relatively short training period.

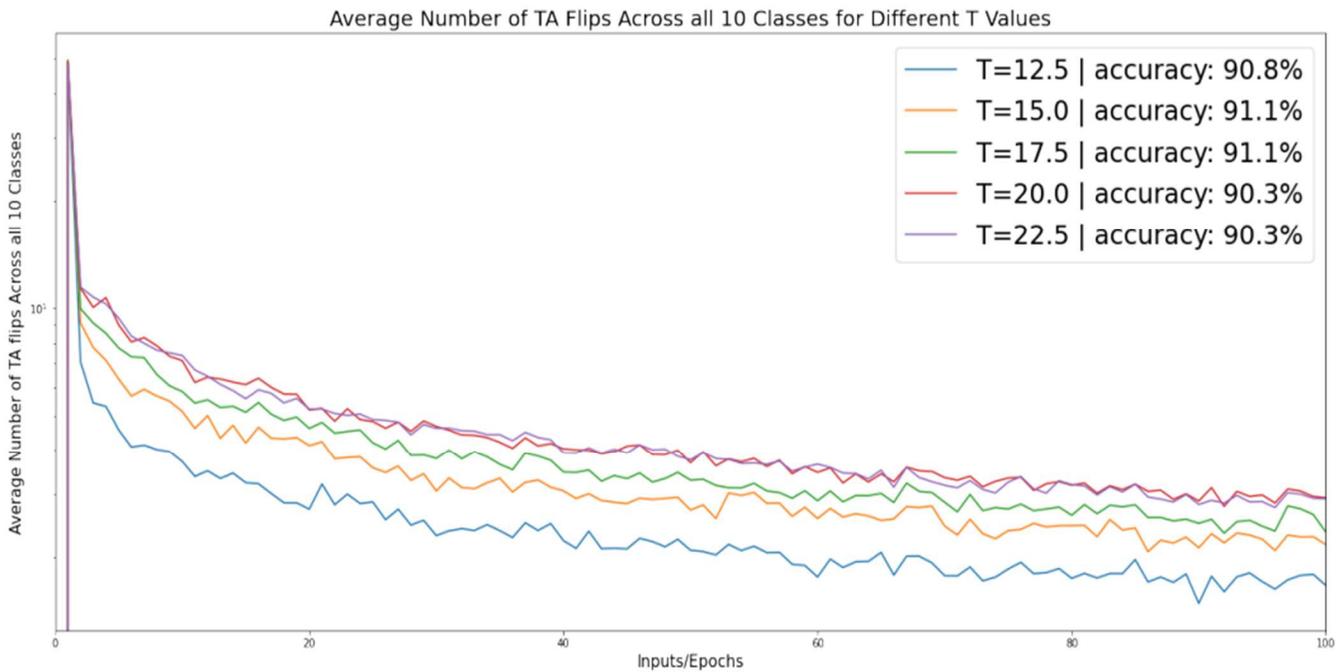
Fig. 6. The typical amount of TA flips for each of the 100 classes for various values of 's'

To identify redundant clauses, the study visualized the frequency of clause outputs over 100 training epochs as shown in Fig. 6. By summing the clause values over time, the study was able to identify clauses that contributed minimally to the TM's decision-making process. For example, in the Noisy XOR dataset, Clause 9 was identified as the least useful for class 0, while Clauses 0, 2, and 4 were identified as least useful for class 1. Although this experiment was conducted on a small dataset, the results indicated that removing redundant clauses could maintain the TM's accuracy while improving interpretability, reducing training time, and enhancing robustness. The study suggested that this method could be further developed for larger datasets, such as MNIST, where a greater number of clauses could be removed over multiple epochs without compromising accuracy.

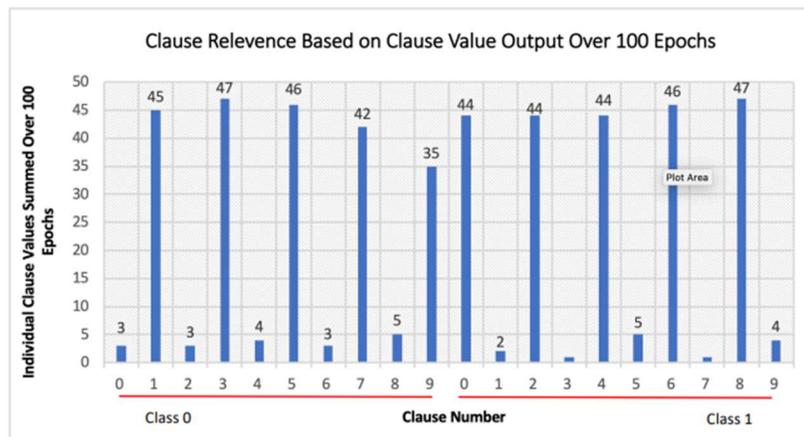
Fig. 7. An illustration displaying the estimated clause significance across 100 epochs

## V. CONCLUSION AND FUTURE WORKS

This study rigorously investigates the visualization of ML, with a specific focus on the TM implementation. We emphasize the critical importance of interpretability and visualization in the study of ML theory, as well as their roles in enhancing the performance of low-complexity architectures. We also provides foundational insights into the theoretical framework of the TM with the results of the visualization efforts, revealing a significant relationship between the learning rate and the number of TA flips. This finding underscores the value of visualization tools in elucidating the inner workings of the TM. The study has largely achieved its objectives. The conclusion posits the implementation of locally stochastic clause dropping as a novel approach, encouraging further exploration of its effects on clause redundancy, thereby indicating a promising direction for future research.